\documentclass[review, times, 10pt]{elsarticle}

\usepackage{amssymb}
\usepackage{caption}

\usepackage{lineno}

\usepackage{tabularx}
\usepackage{adjustbox}
\usepackage{algorithmic}
\usepackage{algorithm}
\usepackage{subfig}
\usepackage{amsmath}
\usepackage{multirow}
\usepackage{color}
\usepackage{xcolor} 

 \usepackage{threeparttable}

 \usepackage[section]{placeins}

\usepackage{natbib}

\begin{document}

\begin{frontmatter}



\title{Inter-object Discriminative Graph Modeling for Indoor Scene Recognition}

\author[label1,label2]{Chuanxin Song}
\author[label1,label2]{Hanbo Wu}
\author[label1,label2]{Xin Ma\corref{mycorrespondingauthor}}
\ead{maxin@sdu.edu.cn}

\affiliation[label1]{organization={Center for Robotics, School of Control Science and Engineering, Shandong University},
            country={China}}
\affiliation[label2]{organization={Engineering Research Center of Intelligent Unmanned System, Ministry of Education},
            country={China}}
\cortext[mycorrespondingauthor]{Corresponding author}

\begin{abstract}
Variable scene layouts and coexisting objects across scenes make indoor scene recognition still a challenging task. Leveraging object information within scenes to enhance the distinguishability of feature representations has emerged as a key approach in this domain. Currently, most object-assisted methods use a separate branch to process object information, combining object and scene features heuristically. However, few of them pay attention to interpretably handle the hidden discriminative knowledge within object information. In this paper, we propose to leverage discriminative object knowledge to enhance scene feature representations. Initially, we capture the object-scene discriminative relationships from a probabilistic perspective, which are transformed into an Inter-Object Discriminative Prototype (IODP). Given the abundant prior knowledge from IODP, we subsequently construct a Discriminative Graph Network (DGN), in which pixel-level scene features are defined as nodes and the discriminative relationships between node features are encoded as edges. DGN aims to incorporate inter-object discriminative knowledge into the image representation through graph convolution and mapping operations (GCN). With the proposed IODP and DGN, we obtain state-of-the-art results on several widely used scene datasets, demonstrating the effectiveness of the proposed approach.
\end{abstract}


\end{frontmatter}


\section{Introduction}
\label{Introduction}
\begin{figure}[htbp]
    \centering
    \includegraphics[width=0.8\textwidth]{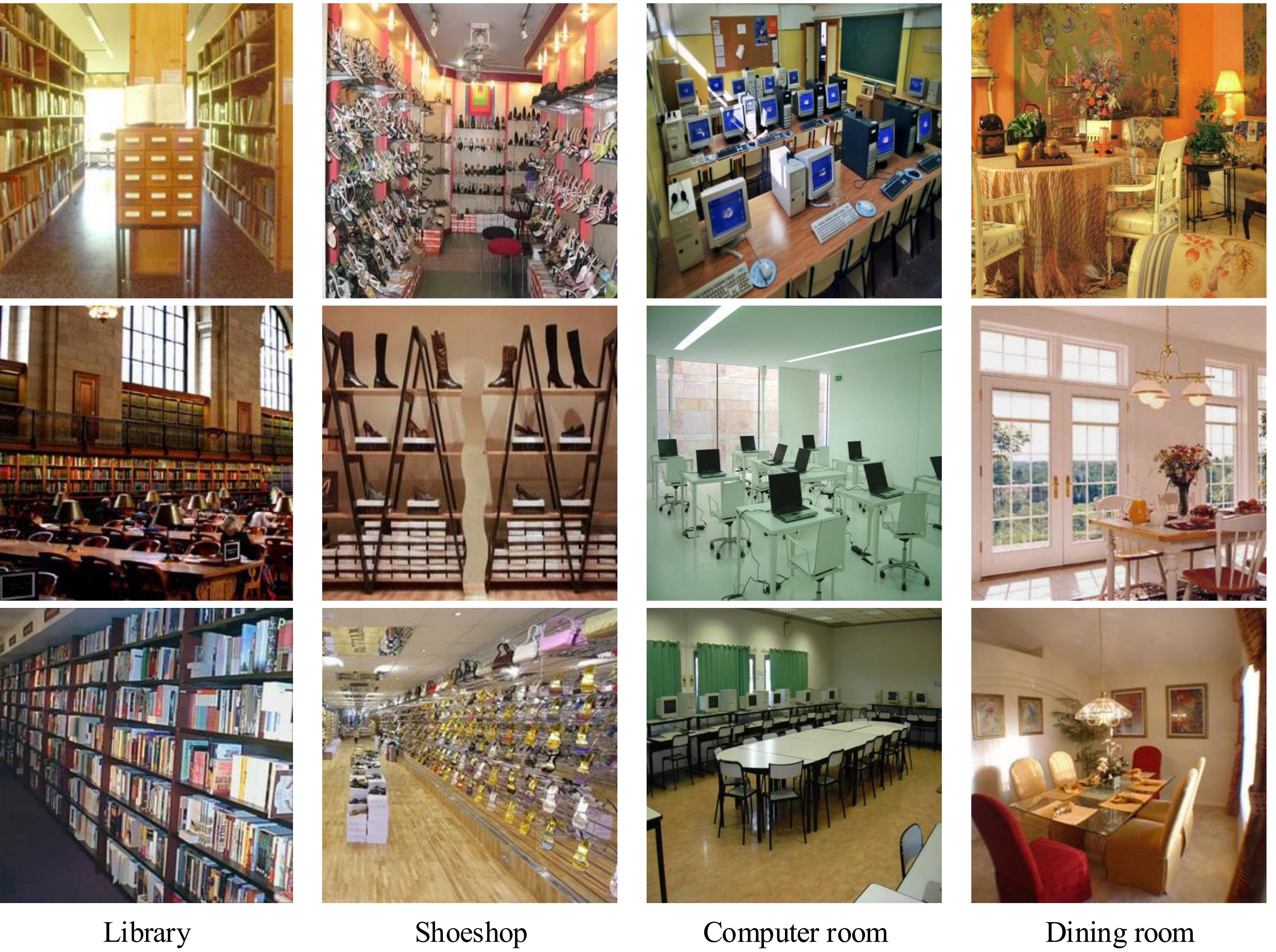}
    \caption{Example images from the MIT-67 dataset. The images in four columns are from four different scene categories ("Library," "Shoeshop," "Computer room," and "Dining room"). It can be seen that the spatial layout within the same category is variable, and the spatial layout within different scenes may be similar (e.g., "Library" and "Shoeshop"); and there may be coexisting objects within different scenes (e.g., "Chair" in scene "Computer room" and "Dining room").}
    \label{Fig_illustration_of_space}
\end{figure}

Scene recognition is a fundamental task in computer vision, and it has attracted significant research attention owing to its diverse applications in domains such as robotics and video analytics \cite{xie2020scener1}. Recognizing indoor scenes from captured images requires exploring not only common characteristics within the same category but also unique characteristics across different categories. However, as shown in Fig. \ref{Fig_illustration_of_space}, the variable scene layouts and coexisting objects across scenes make these characteristics elusive. Consequently, how to extract discriminative image representations remains a central issue in indoor scene recognition.

\begin{figure}[htbp]
    \centering
    \includegraphics[width=\textwidth]{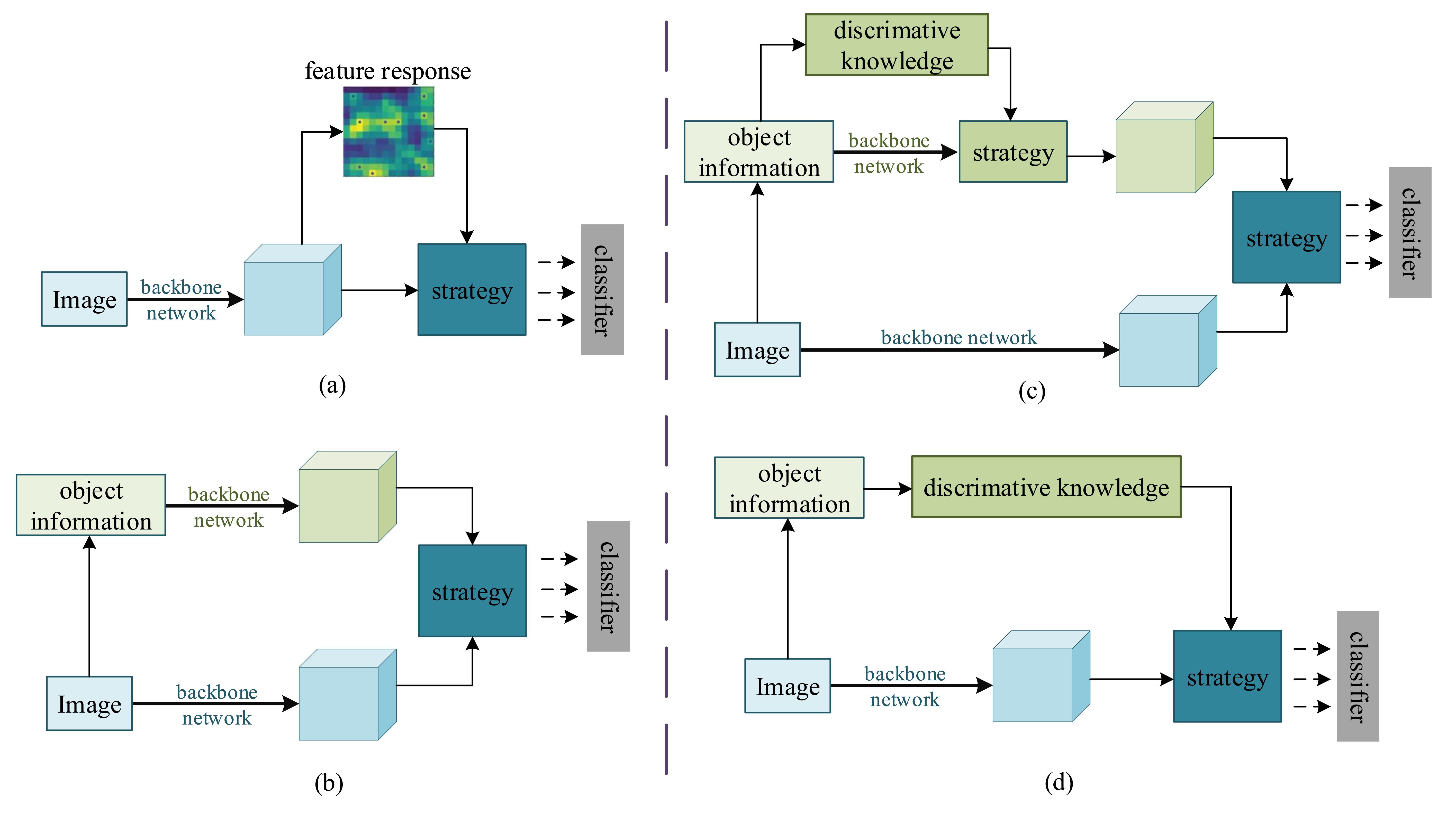}
    \caption{Illustrations of existing and proposed methods. (a) feature response-based method \cite{yuan2019acmr2,chen2020scener3,zhao2018volcanor5,lin2022scener4}, (b) separate object branch-based method (without knowledge) \cite{lopez2020semanticr9,song2023Srrmr10,sceneessencer11,sitaula2021contentr40} and (c) separate object branch-based method (with knowledge) \cite{cheng2018scener7,pereira2021deepr13,zhou2021bormr14,choe2021indoorr16}, and (d) the proposed method.}
    \label{Fig_iillustration_of_existing_method}
\end{figure}

To enhance the discriminability of extracted features, several studies have used pre-trained Convolutional Neural Networks (CNNs) for feature selection. As shown in Fig. \ref{Fig_iillustration_of_existing_method}(a), given an image, they extract representative features by analyzing convolutional channel responses \cite{yuan2019acmr2,chen2020scener3,zhao2018volcanor5} or class activation maps \cite{lin2022scener4}. Subsequently, scene recognition is performed by analyzing the relationships among these salient features. These methods are extremely demanding on the feature discriminability of the pre-trained networks. However, the intrinsic uncertainty of neural networks implies that the high-response regions of CNNs \cite{zhao2018volcanor5} could be focused on non-discriminative regions. This black-box nature of neural networks drives researchers to leverage intra-scene specific object information to obtain more discriminative image representations. 

To integrate object information for scene recognition, some researches \cite{lopez2020semanticr9,song2023Srrmr10,sceneessencer11,sitaula2021contentr40}, as shown in Fig. \ref{Fig_iillustration_of_existing_method}(b), create a separate network branch to extract deep features related to object information. The object-related features are fused with scene features to yield more discriminative representations.  Additionally, as shown in Fig. \ref{Fig_iillustration_of_existing_method}(c), certain methods \cite{cheng2018scener7,pereira2021deepr13,zhou2021bormr14,choe2021indoorr16} attempt to extract discriminative object knowledge from a probabilistic perspective, they then use the discriminative knowledge to adjust the object feature extraction process. These two methods rely on multi-branch architectures to handle the scene image and object information separately. Within this architecture, scene features and object features are fused through concatenation or summation. These fusion methods are entirely heuristic, which may result in suboptimal integration of object knowledge with scene features. 

To alleviate the above limitations, some recent approaches \cite{miao2021objectr12,zeng2020amorphousr15} assign object semantics to scene features, enabling scene recognition through the exploration of contextual relationships among object regions. These approaches provide a more interpretable use of object information and have demonstrated impressive performance.  However, a concern arises regarding the potential weakening of valuable information stored in pixels during the construction of initial features for object regions \cite{yang2023sagnr6}. Moreover, these approaches tend to treat each regional feature equally and overlook the consideration of discriminative object knowledge. A recent study on scene essence \cite{sceneessencer11} emphasizes the connection between scene recognition and the discriminative objects within scenes. Discriminative objects tend to appear in specific scenes, while common objects exhibit similar occurrence probabilities across scenes \cite{cheng2018scener7}. Inspired by this, we argue that features associated with discriminative objects deserve more attention. In line with this, as shown in Fig. \ref{Fig_iillustration_of_existing_method}(d), we propose to map the discriminative object knowledge to feature correlations, aiming at fine-tuning pixel-level scene features from a theoretical perspective, thereby generating more discriminative image representations.

In this paper, we employ statistical methods to capture the object-scene discriminative relationships, which are then converted into inter-object discriminative correlations from a probabilistic perspective. These inferred correlations are quantified within the Inter-Object Discriminative Prototype (IODP), which can provide abundant discriminative object knowledge to assist scene recognition. We frame the discriminative assistance process as a graph representation learning problem, motivated by the graph's flexibility and diversity. Specifically, we propose a Discriminative Graph Network (DGN), in which pixel-level scene features are defined as graph nodes, preserving the entirety of the pixel knowledge. Leveraging IODP, we determine the discriminative relationships among these nodes based on the object semantics they pertain to, which are encoded as graph edges. With the help of Graph Convolution Network (GCN) \cite{kipf2016semir17}, DGN enables the transformation of inter-object discriminative knowledge and scene features into discriminative image representations.

The primary contributions of this paper are listed as follows:

\begin{enumerate}
    \item We propose an Inter-Object Discriminative Prototype (IODP) that contains abundant discriminative object knowledge, which can be used to improve scene recognition performance.
    \item We propose a Discriminative Graph Network (DGN), which leverages the inter-object discriminative prior knowledge from IODP to establish pixel-level correlations within scene features. With the help of GCN \cite{kipf2016semir17}, the scene features are transformed into discriminative image representations.
    \item Extensive experiments conducted on several publicly available scene datasets (MIT-67 \cite{quattoni2009recognizingr18}, SUN397 \cite{xiao2010sunr19}, and Places \cite{zhou2017placesr20}) demonstrate the effectiveness of the proposed approach.
\end{enumerate}

The remainder of this paper is organized as follows. The related works will be described in Section \ref{Related works}. Section \ref{our method} presents the details of the proposed IODP and DGN. Section \ref{Experimental Results and Discussion} describes the experimental details, results, and analysis. Section \ref{conclusion} offers a brief conclusion.

\section{Related works}
\label{Related works}
This section briefly introduces previous studies related to our methods.

\subsection{General deep scene recognition methods}
With the booming development of deep learning, Deep Neural Networks (DNNs) have facilitated the advancement of computer vision \cite{xie2020scener1,liu2022convnetr25}. Accordingly, DNNs-based methods have dominated the scene recognition field \cite{liu2019novelr31,leng2023multitask,yuan2022scaler32}. For instance, Liu et al. \cite{liu2019novelr31} proposed a transfer learning approach based on the ResNet architecture, incorporating multi-layer feature fusion and data augmentation techniques for scene classification. \cite{yuan2022scaler32} presented an efficient Scale Attention module, optimizing features' re-calibration and refinement process to improve scene recognition. Neural networks' powerful feature extraction capability enables these DNNs-based methods to achieve impressive performance. However, in indoor scene recognition, the performance improvement of deep learning is not as remarkable as in fields like image classification \cite{lopez2020semanticr9,song2023Srrmr10}. This phenomenon is primarily because different indoor scene categories present similar features, which stem from the intricate scene layouts and the coexisting objects across scenes.

For better scene recognition, many researchers \cite{yuan2019acmr2,chen2020scener3,zhao2018volcanor5,lin2022scener4} tend to extract more discriminative image representations to distinguish scenes. Adi-Red \cite{zhao2018volcanor5} employs a neural network pretrained on large datasets to locate salient image regions. These salient regions are then used to extract discriminative feature representations through multiple network branches. Similarly, \cite{chen2020scener3} proposes using convolutional activation maps generated by CNNs to identify discriminative regions, integrating them using Graph Convolution Networks. Also, the authors of MRNet \cite{lin2022scener4} design a salient local feature extractor based on CAM \cite{zhou2016learningr34}. They reuse one CNN to generate class activation maps and extract local distinctive features. As shown in Fig. \ref{Fig_iillustration_of_existing_method} (a), a common characteristic among these studies is their substantial dependence on the feature discriminator of the pre-trained network. However, neural networks' inherent uncertainty implies that CNNs' high-response regions may focus on non-discriminative regions. In contrast, this paper emphasizes leveraging specific object information within scenes to assist in obtaining discriminative features, avoiding potential limitations associated with the black-box nature of neural networks.

\subsection{Object information-guided scene recognition methods}
Given the significance of intra-scene object information for scene recognition, incorporating object information to improve the discrimination of image representations has evolved into a mainstream approach. As shown in Fig. \ref{Fig_iillustration_of_existing_method} (b), some methods \cite{lopez2020semanticr9,song2023Srrmr10,sceneessencer11,sitaula2021contentr40,pal2019deducer37} directly leverage object information within scenes in an end-to-end manner. For instance, SRRM \cite{song2023Srrmr10} introduced an Adaptive Confidence Filter to mitigate the negative impact of object semantic ambiguity and generated a deep representation of object information through a separate network. \cite{sceneessencer11} introduced a graph neural network to explore the relationship between object information within scenes, facilitating the generation of object-level discriminative features. These methods create a dedicated branch to extract object-related features end-to-end.

Furthermore, some studies \cite{cheng2018scener7,pereira2021deepr13,zhou2021bormr14,choe2021indoorr16}, as shown in Fig. \ref{Fig_iillustration_of_existing_method} (c), contend that latent knowledge hidden in object information can be harnessed to support scene recognition. SDO \cite{cheng2018scener7} captured the statistical attributes of objects to exploit the correlation among objects across scenes, improving inter-class discriminability by selecting representative objects. Also, a semantic inter-object relationship approach was proposed by \cite{pereira2021deepr13}, which exploits object knowledge within scenes based on the distance relations between objects. These methods extract object prior knowledge from a probabilistic perspective. However, the prior knowledge is only used for object information processing in these methods. They still use a multi-branch architecture to process object information. In this architecture, scene features and object features can only be fused through concatenation or summation. Such heuristic fusion methods potentially lead to suboptimal combinations of object knowledge and scene features.

To enhance the integration of object information with scene features, some methods \cite{miao2021objectr12,zeng2020amorphousr15,yang2023sagnr6} assign object semantics to scene features for more intuitive use of object information. For instance, OTS-Net \cite{miao2021objectr12} assigned object semantics to segmentation network features and established relationships between semantic features through the attention mechanism. This approach leverages segmentation networks for scene recognition, ensuring a lightweight solution. However, it does not address the disparity between pre-trained models of segmentation networks and recognition networks. Another approach, ARG-Net \cite{zeng2020amorphousr15}, assigned object semantics to scene features pre-trained on scene datasets. This approach uses object information to guide the aggregation of scene feature regions to attain discerning representations. The feature aggregation process in these methods is based on object guidance, making the utilization of object information more transparent. Nevertheless, these methods are constrained to heuristic uses of object information, neglecting the exploitation of latent knowledge within object information. 

In this paper, we also assign object semantics to scene features. Unlike the methods above, our approach aims to utilize the discriminative prior knowledge among objects to guide the modification of scene features, resulting in improved discriminability of feature representations.

\subsection{Graph convolution networks}
Given the excellent ability to handle diverse topologies and analyze unstructured data, Graph Convolution Networks (GCN) \cite{kipf2016semir17} has been widely used in various fields, including action recognition, pedestrian trajectory prediction, sentiment analysis, and more \cite{yang2023sagnr6,tian2022skeletonr28,zhou2023static,tian2023skeletonr29,liu2023enhancingr30}. For action recognition, Tian et al. \cite{tian2022skeletonr28} introduced an attention-enhanced spatial-temporal graph convolution network, which boosts the learning of fine-grained abnormal gait features. For pedestrian trajectory prediction, a data-driven static and dynamic graph encoding network was proposed by \cite{zhou2023static}, explicitly modeling the social interactions among pedestrians for trajectory prediction. As for sentiment analysis, the authors of \cite{liu2023enhancingr30} presented a dual-gated GCN that leverages contextual affective knowledge to conduct aspect-based sentiment analysis.

Building on the success of Graph demonstrated in previous studies, this paper frames our discriminative assistance process as a graph representation learning problem. Due to the intricate scene composition, several studies \cite{chen2020scener3,zeng2020amorphousr15,zhou2023attentionalr38,zeng2023naturalScene}] have also used GCN to explore advanced image representations for scene classification. Chen et al. \cite{chen2020scener3} use convolutional activation mapping to identify discriminative regions within scenes. ARG-Net \cite{zeng2020amorphousr15} defines the regional representations as nodes using semantic segmentation. These methods then use GCN to explore the contextual relations between regional features. Nonetheless, they employ superpixels as nodes to construct region-level features, which could weaken the valuable information stored in pixels \cite{yang2023sagnr6}. Moreover, these approaches tend to treat each regional feature equally and ignore the use of discriminative object knowledge. In contrast, this paper encodes the discriminative correlations between objects as edges and adjusts pixel-level scene features using GCN. In this way, the feature's original information and discriminative object knowledge are fully utilized to obtain discriminative scene representations.

\section{Our method}
\label{our method}
\begin{figure}[htbp]
    \centering
    \includegraphics[width=0.8\textwidth]{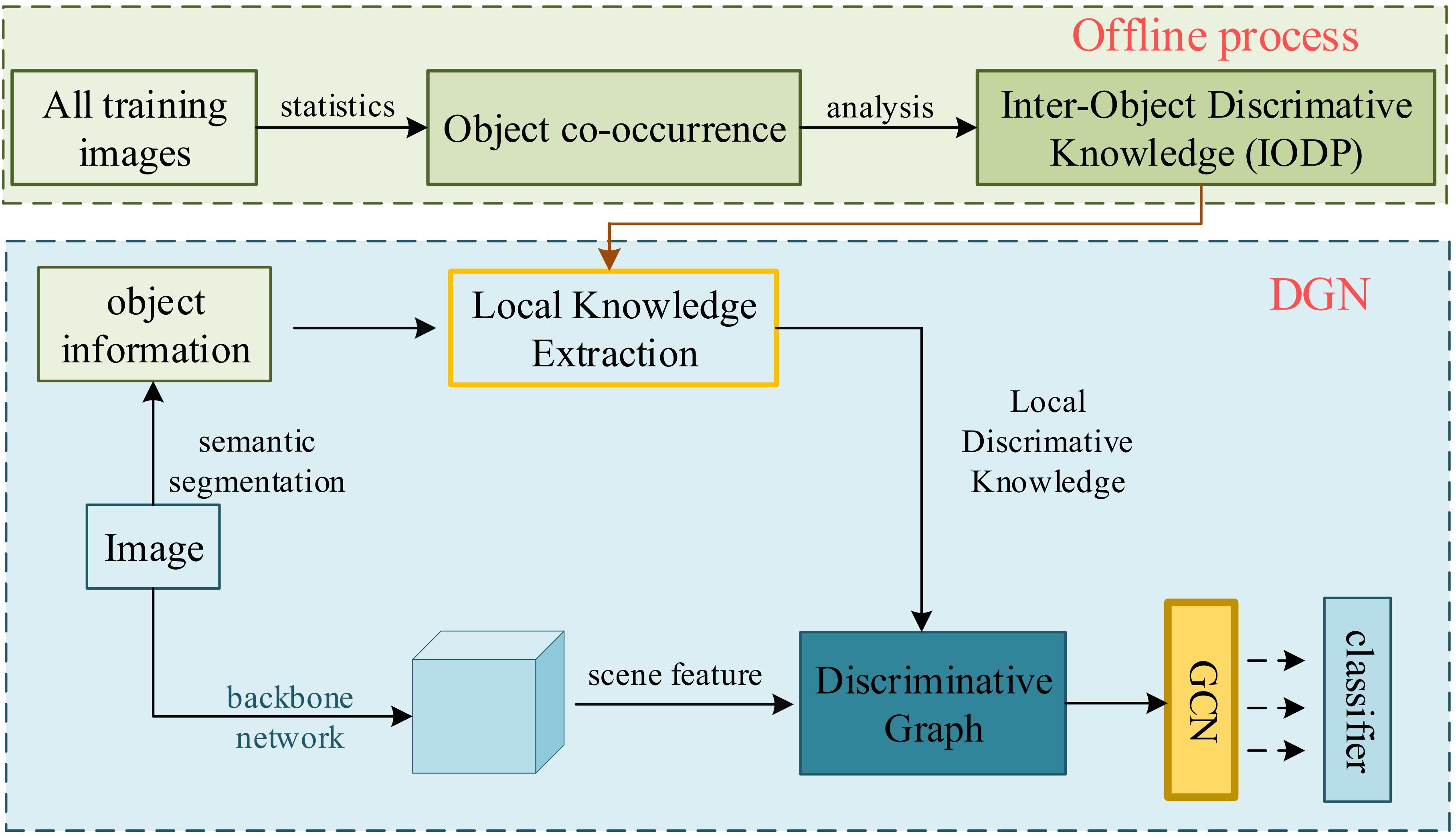}
    \caption{Overall workflow of the proposed approach for scene recognition. We first statistics and analyze all the training data to get the Inter-Object Discriminative Prototype (IODP). Then, we build a Discriminative Graph Network (DGN) upon IODP to integrate discriminative object knowledge into image representations.}
    \label{Fig_Overall_process}
\end{figure}

In this section, we first elucidate the motivations of the proposed approach and then introduce the construction of Inter-object Discriminative Prototype (IODP) from training data. Subsequently, we build a Discriminative Graph Network (DGN) upon IODP to integrate discriminative object knowledge into image representations. The overall conceptual workflow of our proposed approach is presented in Fig. \ref{Fig_Overall_process}; we will describe it in detail next.

\subsection{Motivation and overview}
Recent researches \cite{sceneessencer11,cheng2018scener7} emphasize the importance of discriminative objects in scene recognition, as these objects are unique to specific scenes, unlike common objects that occur with similar frequency across scenes. This insight motivates us to pay more attention to the scene features associated with these discriminative objects to obtain a more discriminative image representation.

The ensuing challenge is quantifying the attention allocated to each object-related feature. Inspired by SDO \cite{cheng2018scener7}, we employ a Bayesian statistical approach to establish the discriminative correlation between objects through an Inter-Object Discriminative Prototype (IODP). In line with this, we link the prior knowledge within IODP with scene features using graph theory. Specifically, we propose a Discriminative Graph Network (DGN), where pixel-level scene features are defined as graph nodes. Leveraging IODP, we determine the discriminative relationships among these nodes based on their object semantics, which are encoded as graph edges. The DGN adjusts each pixel-level scene feature through graph convolution and mapping operations (GCN \cite{kipf2016semir17}), ultimately yielding a more discerning image representation.

\subsection{Inter-object discriminative prototype}
\label{Inter-Object Discriminative Prototype}
\begin{figure}[htbp]
    \centering
    \includegraphics[width=\textwidth]{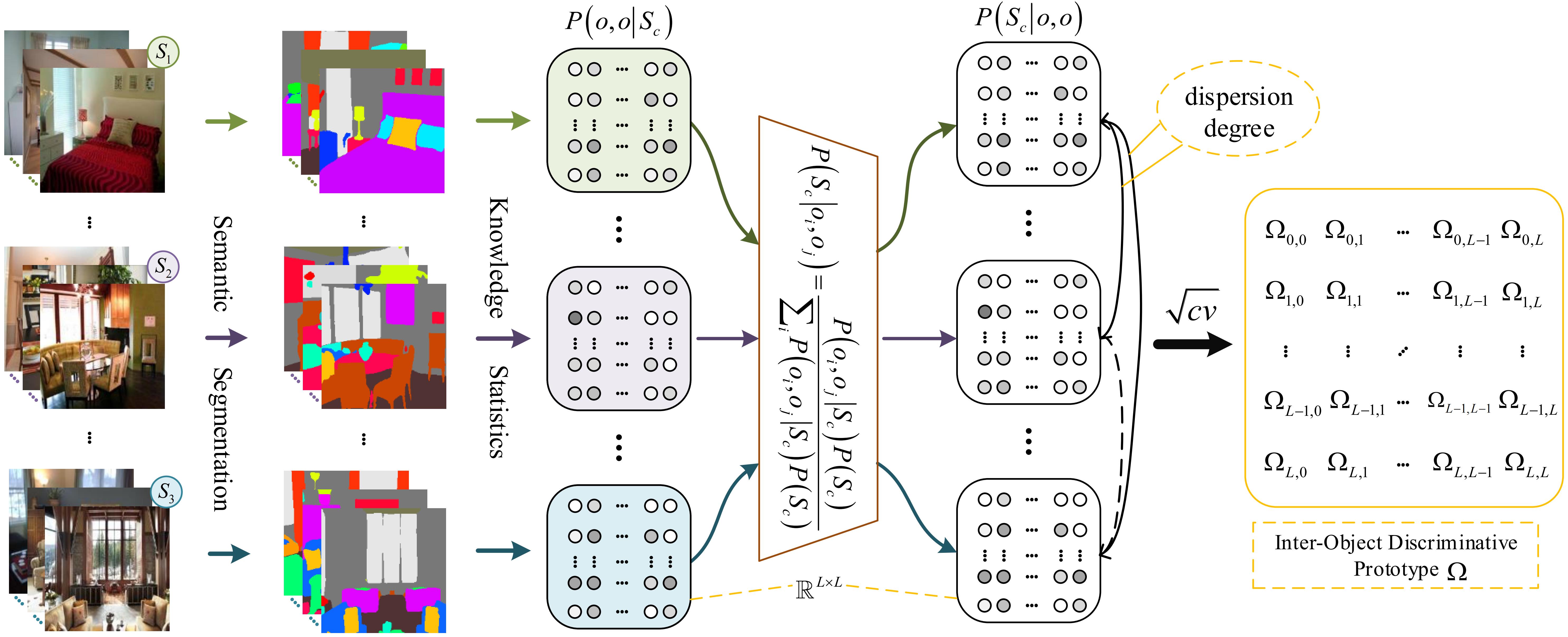}
    \caption{A description of the process of constructing the Inter-Object Discriminative Prototype (IODP). The whole process is based on the training data from scene datasets. Different colored blocks represent different scene categories, where $S_c$ denotes the $c_{th}$ scene category. The resulting IODP appears as a matrix with $L \times L$ dimensions. $L$ denotes the number of segmentable objects by the semantic segmentation technique. $\Omega_{i,j}$} signifies the discriminative correlation between object $o_i$ and $o_j$.
    \label{Fig1}
\end{figure}

Drawing inspiration from SDO\cite{cheng2018scener7}, we employ Bayesian probabilistic statistics to delve into object knowledge. The focus of SDO is on the discrimination of individual objects. Given that our ultimate goal is to uncover discriminative correlations within scene features, we shift our focus to exploring inter-object discriminative knowledge. Accordingly, we propose an Inter-Object Discriminative Prototype (IODP), as illustrated in Fig. \ref{Fig1}.

In Fig. \ref{Fig1}, we first determine the distribution of object pairs across various scenes using semantic segmentation and statistical analysis on the training data from scene datasets. Given a scene dataset with $C$ scene categories, we denote the $c_{th}$ scene category of it by $S_c$ and $c \in \left\{ {1,2,...,C} \right\}$. Upon processing each instance in $S_c$, we can compute the conditional probability $P\left( {{o_i},{o_j}\left| {{S_c}} \right.} \right)$ of encountering objects $o_i$ and $o_j$ together within $S_c$. Specifically, for Places\_7 and Places\_14 \cite{zhou2017placesr20} datasets that have sufficient data, we calculate this probability directly. However, for the MIT-67 \cite{quattoni2009recognizingr18} and SUN397 \cite{xiao2010sunr19} datasets, the relatively limited data volume poses challenges to data representativeness. Therefore, we compute this probability by assuming object distribution independence, i.e., we calculate the distributions of $o_i$ and $o_j$ separately, and then combine them to approximate the joint distribution $P\left( {{o_i},{o_j}\left| {{S_c}} \right.} \right)$, reducing potential errors. Specific details are formulated in Eq. \ref{eq1}:

\begin{equation}
P\left( {{o_i},{o_j}\left| {{S_c}} \right.} \right) = \left\{ {\begin{array}{*{20}{c}}
  {\frac{{{N_{{o_i},{o_j}}}}}{{{N_{{S_c}}}}},}&{Places\_7\ or\ Places\_14} \\ 
  {P\left( {{o_i}\left| {{S_c}} \right.} \right)P\left( {{o_j}\left| {{S_c}} \right.} \right) = \frac{{{N_{{o_i}}} \times {N_{{o_j}}}}}{{N_{{S_c}}^2}},}&{MIT-67\ or\ SUN397} 
\end{array}} \right.
\label{eq1}
\end{equation}
where $N_{S_c}$ denotes the total number of instances in scene category $S_c$. For any objects $o_i$ and $o_j$, ${N_{{o_i},{o_j}}}$ denotes the number of instances in which they occur simultaneously. ${N_{{o_i}}}$ and ${N_{{o_j}}}$ respectively refer to the number of instances in which $o_i$ and $o_j$ appear. The values of $i$ and $j$ fall within the range of $[1, L]$ where $L$ denotes the number of object categories segmentable by the semantic segmentation technique. The comparative performance of these two computational approaches is discussed in Section \ref{Object Co-occurrence Calculations}.

Once all object co-occurrence distributions are computed, $P\left( {\left. {{S_c}} \right|{o_i},{o_j}} \right)$ can be deduced through the Bayesian posterior probability formulation:
\begin{equation}
    P\left( {\left. {{S_c}} \right|{o_i},{o_j}} \right) = \frac{{P\left( {{o_i},{o_j}\left| {{S_c}} \right.} \right)P\left( {{S_c}} \right)}}{{\sum\nolimits_c {P\left( {{o_i},{o_j}\left| {{S_c}} \right.} \right)P\left( {{S_c}} \right)} }}
    \label{eq2}
\end{equation}
where $P\left( {{S_c}} \right)$ represents the probability of scene category $S_c$, $P\left( {{S_c}} \right) = 1/C$.

Subsequently, given the presence of $o_i$, $o_j$ in scenes, we can construct the posterior probability matrix $P\left( {\left. S \right|{o_i},{o_j}} \right)$  for all $C$ scene categories:
\begin{equation}
    P\left(S \mid o_{i}, o_{j}\right)=\left[P\left(S_{1} \mid o_{i}, o_{j}\right), \ldots, P\left(S_{c} \mid o_{i}, o_{j}\right), \ldots, P\left(S_{C} \mid o_{i}, o_{j}\right)\right]
    \label{eq3}
\end{equation}

\begin{figure}[htbp]
\centering
\subfloat[]{\includegraphics[width=5cm]{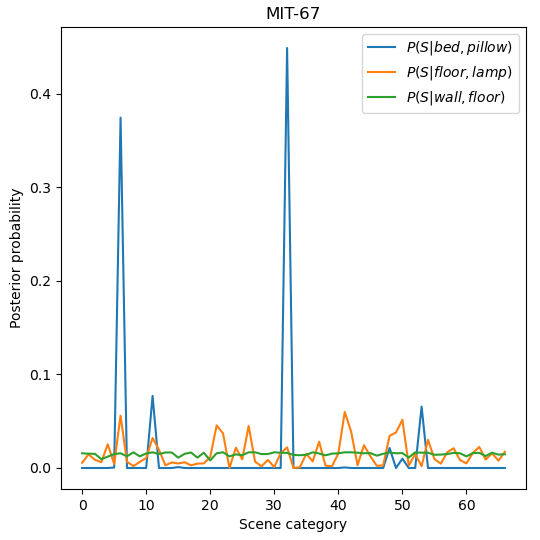}%
\label{Fig_2_1_case}}
\hfil
\subfloat[]{\includegraphics[width=5cm]{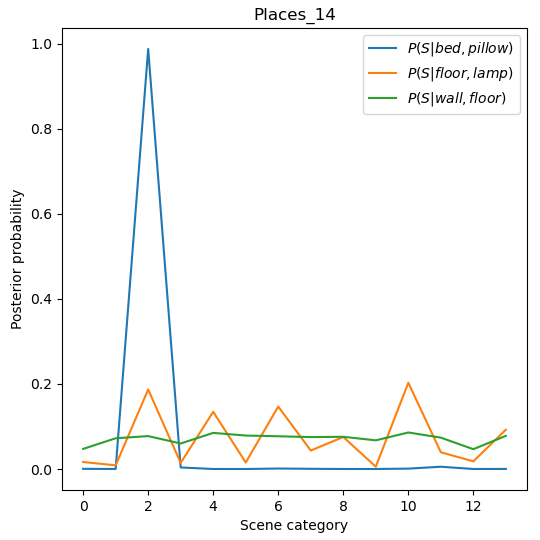}%
\label{Fig2_2_case}}
\caption{Posterior probabilities Visualization for different scene categories when given several object pairs in MIT-67 and Places\_14 datasets.}
\label{Fig2}
\end{figure}

We use Fig. \ref{Fig2} to visualize the posterior probability matrices of the object pairs \{bed, pillow\}, \{floor, lamp\}, and \{wall, floor\} for different scene categories. It can be seen that \{bed, pillow\} exhibits high probabilities in specific scenes and low in others, indicating a substantial discriminative significance. Conversely, \{wall, floor\} displays similar probabilities for all scenes, offering minimal assistance in scene recognition. The discriminative significance of \{floor, lamp\} lies between \{bed, pillow\} and \{wall, floor\}. To summarize, the greater the fluctuation in the distribution of the posterior probability matrix, the more discriminative the object pair. Therefore, we propose to use the dispersion degree of $P\left( {\left. S \right|{o_i},{o_j}} \right)$ to measure the discriminative correlation between $o_i$ and $o_j$. We apply several common metrics to measure dispersion, including Range ($R$), Standard deviation ($\sigma$), and Coefficient of variation ($cv$):
\begin{equation}
    {\theta _{i,j}} = \left\{ {\begin{array}{*{20}{c}}
  {\max \left( {P\left( {\left. S \right|{o_i},{o_j}} \right)} \right) - \min \left( {P\left( {\left. S \right|{o_i},{o_j}} \right)} \right),R} \\ 
  {\sigma \left( {P\left( {\left. S \right|{o_i},{o_j}} \right)} \right),\sigma } \\ 
  {cv\left( {P\left( {\left. S \right|{o_i},{o_j}} \right)} \right) = \frac{{\sigma \left( {P\left( {\left. S \right|{o_i},{o_j}} \right)} \right)}}{{\mu \left( {P\left( {\left. S \right|{o_i},{o_j}} \right)} \right)}},cv} 
\end{array}} \right.
\label{eq4}
\end{equation}
where ${\theta _{i,j}}$ denotes the dispersion degree of $P\left( {\left. S \right|{o_i},{o_j}} \right)$. $\sigma (\cdot)$ is the Standard Deviation of $P\left( {\left. S \right|{o_i},{o_j}} \right)$ and $\mu (\cdot)$ is the Arithmetic Mean of $P\left( {\left. S \right|{o_i},{o_j}} \right)$.

\begin{algorithm}[htbp]
\caption{Inter-Object Discriminative Prototype algorithm}\label{alg:alg1}
\textbf{Input}: Training instances of $C$ scene categories: $S_c, c=1,...C$;\\
\textbf{Output}: Inter-Object Discriminative Prototype $\Omega \in {\mathbb{R}^{L \times L}} $;
\begin{algorithmic}[1] 
\STATE \COMMENT{$L$ represents the number of object categories} 
\FOR{$c=1$ to $C$}
    \STATE Extract the semantic segmentation label map for all instances within $S_c$;
    \FOR{$i=1$ to $L$}
        \FOR{$j=1$ to $L$}
        \STATE Process all label maps, obtain object co-occurrence distributions $P\left( {{o_i},{o_j}\left| {{S_c}} \right.} \right)$;
        \STATE $P\left(S_{c} \mid o_{i}, o_{j}\right)=\frac{P\left(o_{i}, o_{j} \mid S_{c}\right) P\left(S_{c}\right)}{\sum_{c} P\left(o_{i}, o_{j} \mid S_{c}\right) P\left(S_{c}\right)}$;
        \ENDFOR
    \ENDFOR
\ENDFOR
\FOR{$i=1$ to $L$}
    \FOR{$j=1$ to $L$}
    \STATE $P\left( S \mid o_i, o_j \right) = \left[ P\left( S_1 \mid o_i, o_j \right), \ldots, P\left( S_c \mid o_i, o_j \right), \ldots, P\left( S_C \mid o_i, o_j \right) \right]$;
    
    \STATE ${\theta_{i,j}} = cv\left( {P\left( {\left. S \right|{o_i},{o_j}} \right)} \right) = \frac{{\sigma \left( {P\left( {\left. S \right|{o_i},{o_j}} \right)} \right)}}{{\mu \left( {P\left( {\left. S \right|{o_i},{o_j}} \right)} \right)}}$;
    \STATE ${\Omega_{i,j}} = \sqrt {{\theta _{i,j}}}$;
    \ENDFOR
\ENDFOR
\STATE \textbf{return} $\Omega$;
\end{algorithmic}
\end{algorithm}

Furthermore, since the above process ignores some details (color, etc.) in the image, which could result in the final Inter-Object Discriminative Prototype $\Omega$ being overly sharp and unrealistic. To prevent this, we passivate ${\theta_{i,j}}$ to obtain the discriminative correlation ${\Omega_{i,j}}$ by:

\begin{equation}
    {\Omega_{i,j}} = \sqrt {{\theta _{i,j}}}
    \label{eq5}
\end{equation}

We have conducted ablation experiments in Section \ref{Inter-Object discrimination correlation measurements} to choose the optimal way to measure inter-object discriminative correlation.

After computing the discriminative correlation for each object pair, we obtain the final Inter-Object Discriminative Prototype $\Omega \in {\mathbb{R}^{L \times L}}$ (IODP), which encompasses rich inter-object discriminative knowledge. The pseudocode for generating IODP is presented in Algorithm \ref{alg:alg1}.

\subsection{Discriminative Graph Network}
\begin{figure}[htbp]
    \centering
    \includegraphics[width=1\textwidth]{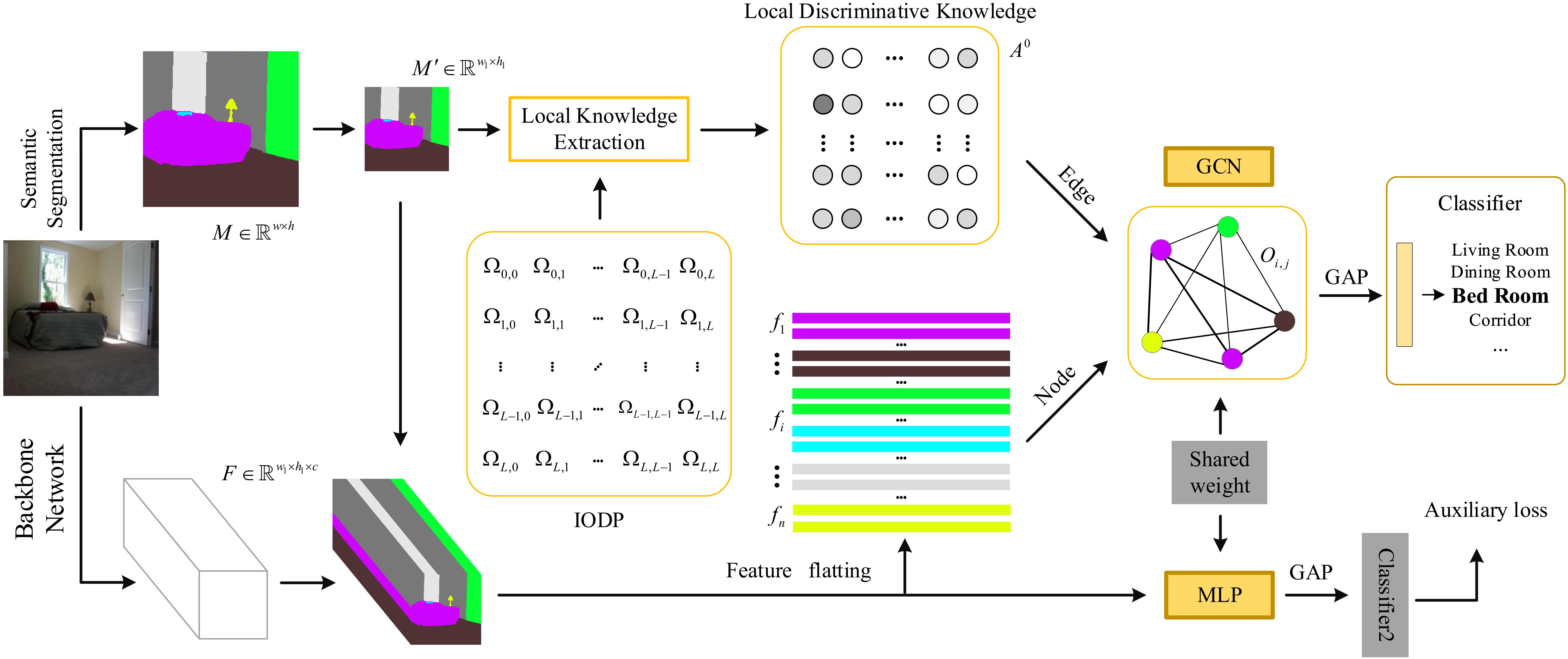}
    \caption{The overall process of the proposed Discriminative Graph Network (DGN). The input image $I$ is initially fed into a pre-trained backbone network to obtain the feature map $F \in {\mathbb{R}^{{w_1} \times {h_1} \times c}}$. Simultaneously, a semantic segmentation network is used to generate the label map $M \in {\mathbb{R}^{w \times h}}$ of $I$. $M$ provides the object semantics corresponding to the pixel-level features within $F$. Subsequently, all pixel-level features are represented as nodes, and the discriminative relationships between these nodes, provided by the IODP, are encoded as edges. Next, GCN is performed to give more attention to features associated with discriminative objects, resulting in discerning image representations. Meanwhile, an auxiliary loss is introduced to reinforce the training of the backbone network parameters. The details for the process of Local Knowledge Extraction can be seen in Algorithm \ref{alg:alg2}.}
    \label{Fig3}
\end{figure}

To leverage the inferred IODP for enhancing the discriminative ability of scene features, we formulate a framework known as the Discriminative Graph Network (DGN). The entire process of the DGN is depicted in Fig. \ref{Fig3}.

\begin{figure}[htbp]
    \centering
    \includegraphics[width=0.8\textwidth]{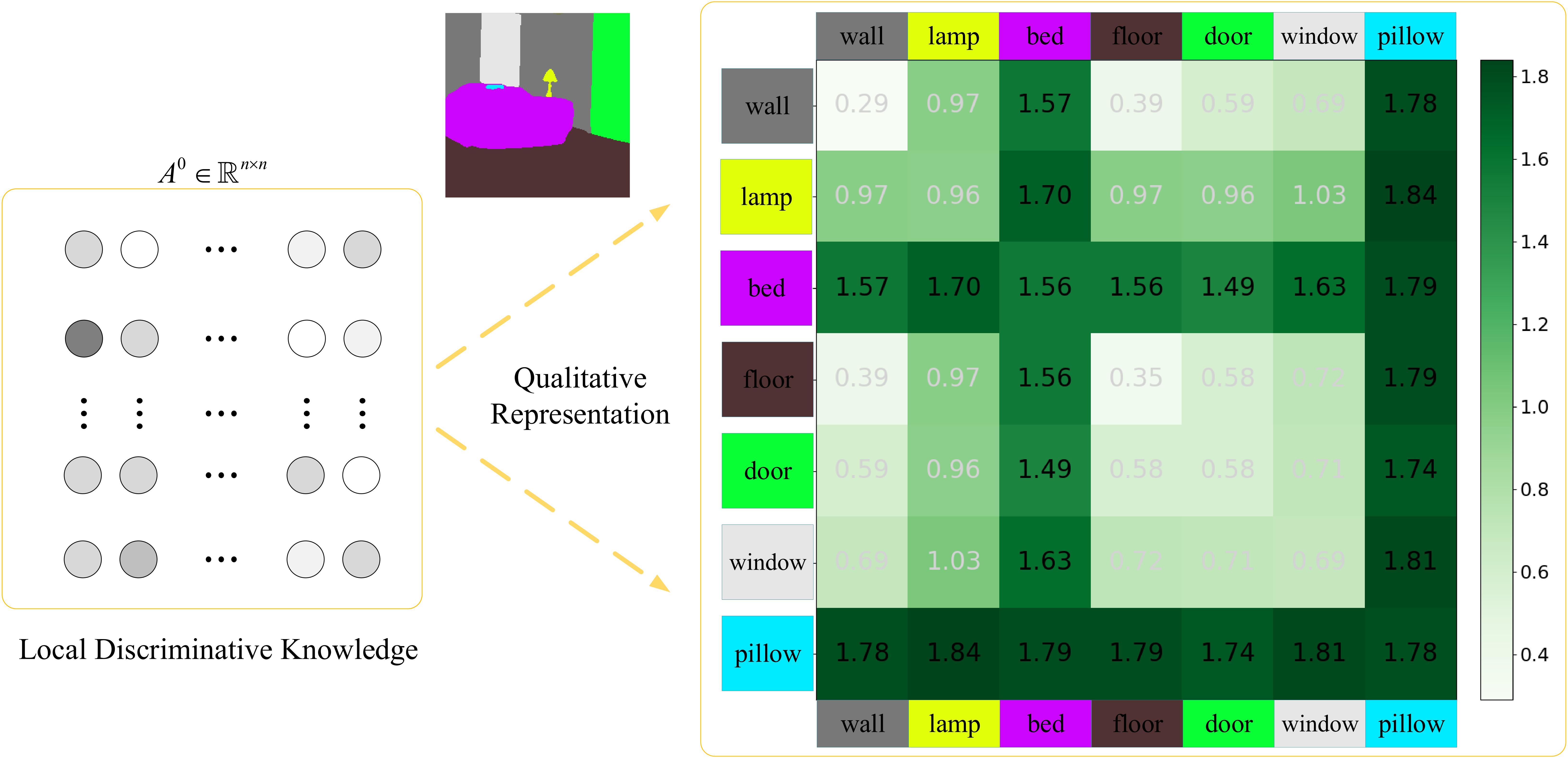}
    \caption{A qualitative representation of the local discriminative knowledge ${A^0}$ generated by IODP for the label map in Fig. \ref{Fig3}. Note that the presented qualitative representation is not ${A^0}$, but all element values in ${A^0}$, i.e., discriminative relationships between node representations can be seen in the presented representation. Darker colors indicate a stronger discrimination between corresponding objects, while lighter colors indicate a weaker discrimination.}
    \label{Fig4}
\end{figure}

\subsubsection{Constructing a discriminative graph}
Given an image $I \in {\mathbb{R}^{w \times h \times 3}}$, we can denote its corresponding semantic segmentation label map as $M \in {\mathbb{R}^{w \times h}}$. Feeding $I$ into the backbone network, we can obtain the corresponding scene feature map $F \in {\mathbb{R}^{{w_1} \times {h_1} \times c}}$. Subsequently, we employ nearest-neighbor interpolation on $M$ to acquire $M' \in {\mathbb{R}^{{w_1} \times {h_1}}}$ with the same resolution as $F$. With $M'$, we can obtain the object semantics corresponding to the pixel-level features within $F$ \cite{miao2021objectr12,zeng2020amorphousr15}. Based on the above representation and the prior knowledge in IODP, we construct a discriminative graph $G = \left( {V,E,A} \right)$, where $V$ denotes the nodes of the graph, $E$ denotes the edges, and the adjacency matrix of the graph is represented as $A$. Unlike previous approaches \cite{chen2020scener3,zeng2020amorphousr15} that require aggregation of region-level features as node representations, which may corrupt the original topological nature of the scene features \cite{yang2023sagnr6}. Thanks to that our IODP has both discriminative knowledge between different objects and between identical objects, we can use all pixel-level features to construct $V$, ensuring that the scene feature information remains untarnished.

Specifically, all feature information within $F$ is used to construct graph nodes $V = [{f_1},{f_2},...,{f_i},...{f_n}]$, where ${f_i} \in {\mathbb{R}^c}$ represents the ${i_{th}}$ pixel-level feature of $F$, and $n = {w_1} \times {h_1}$. Edges $E$ between nodes are encoded as discriminative relationships between nodes' corresponding semantic objects, which can be obtained from our IODP. For ease of processing, we convert $M'$ to the one-dimensional $M'' = [{m_1},{m_2},...,{m_i},...,{m_n}]$ corresponding to the nodes in $V$. Subsequently, the local discriminative knowledge ${A^0} \in {\mathbb{R}^{n \times n}}$ among $n$ nodes can be extracted by Algorithm \ref{alg:alg2}. Taking the input image in Fig. \ref{Fig3} as an example, the qualitative representation about ${A^0}$ is illustrated in Fig. \ref{Fig4}.

\begin{algorithm}[htbp]
\caption{Algorithm for extracting local discriminative knowledge}\label{alg:alg2}
\textbf{Input}: $V = [{f_1},{f_2},...,{f_i},...{f_n}]$, IODP $\Omega$, $M'' = [{m_1},{m_2},...,{m_i},...,{m_n}]$;\\
\textbf{Output}: Local Discriminative Knowledge  ${A^0} \in {\mathbb{R}^{n \times n}}$;
\begin{algorithmic}[1] 
\FOR{$i=1$ to $n$}
    \STATE Obtain the object semantics ${m_i}$ corresponding to ${f_i}$ from $M''$;
    \FOR{$j=1$ to $n$}
        \STATE Obtain the object semantics ${m_j}$ corresponding to ${f_j}$ from $M''$;
        \STATE $A_{i,j}^0 = \Omega_{{m_i},{m_j}}$;
    \ENDFOR
\ENDFOR
\STATE \textbf{return} $A^0$;
\end{algorithmic}
\end{algorithm}

As shown in Fig. \ref{Fig4}, the inter-node discriminative values in ${A^0}$  fall outside the interval [0, 1], which is not conducive to subsequent processing. Consequently, we perform row normalization on ${A^0}$ to generate the adjacency matrix $A$ for our discriminative graph:
\begin{equation}
    {A_{i,j}} = A_{i,j}^0/sum({A_{i,:}})
    \label{eq6}
\end{equation}

Following the above procedure, we have successfully constructed our discriminative graph.

\subsubsection{GCN}
In the previous steps, we constructed our discriminative graph $G = \left( {V,E,A} \right)$, where $V \in {\mathbb{R}^{n \times c}}$ denotes the node set and $A \in {\mathbb{R}^{n \times n}}$ is the corresponding adjacency matrix. Considering the proven effectiveness of GCN \cite{kipf2016semir17} for graph optimization, we apply it to our discriminative graph to update graph nodes and obtain an optimal image representation. The process of applying GCN can be formulated by:
\begin{equation}
\begin{gathered}
  V^* = \eta ({D^{ - 1}}\widetilde AVW) \hfill \\
  \widetilde A = A + {I_N} \hfill \\ 
\end{gathered} 
\label{eq7}
\end{equation}
where ${V^*}$ represents the optimized node representations, $W \in {\mathbb{R}^{c \times d}}$ is the trainable weight matrix, $d$ is the hidden dimension of GCN. ${D_{i,i}} = \sum\nolimits_j {{{\widetilde A}_{i,j}}}$ is the degree matrix of $\widetilde A$, and $\eta$ denotes the activation function $Sigmoid$.

To prevent overfitting \cite{zeng2020amorphousr15}, we only adopt one GC layer to obtain the optimized node representation ${V^*} \in {\mathbb{R}^{n \times d}}$.

\subsubsection{Scene classifier and auxiliary loss}
\label{Scene Classifier and Auxiliary loss}
Applying Global Average Pooling (GAP) on ${V^*}$, we can obtain the optimized image representation $f^o$. $f^o$ is fed into a fully connected network for scene category prediction. During training, the softmax and cross-entropy function are used as the loss function to compute the classification loss $l_o$.

As discussed in the preceding sections, we employ a discriminative graph to incorporate the inter-object prior knowledge from IODP into scene features, improving the discrimination of final image representations. However, this process raises an underlying issue: the potential for suboptimal adjustment of the backbone network parameters during training. Specifically, the DGN leverages the inter-object prior knowledge from IODP to improve the discrimination of scene features, resulting in a diminished model loss. Such diminution may inadvertently lead to suboptimal optimization of the backbone network parameters. To mitigate this concern, we introduce an auxiliary classifier within the DGN. This classifier generates a dedicated auxiliary loss that specifically targets the refinement of the backbone network parameters, thereby facilitating their fine-tuning.

Specifically, the node feature representation $V$ is not only fed into the GCN layer but also into a Multi-Layer Perceptron (MLP) layer. The weights of this MLP layer are shared with the $W$ of the hidden layer of the GCN layer to ensure consistent parameters. We then add an auxiliary scene classifier after the MLP layer, which generates an auxiliary loss \(l_a\). In this way, we leverage the prior knowledge provided by our IODP to enhance feature discrimination, while ensuring adequate training of the backbone parameters. Note that the weights of the auxiliary scene classifier is independent of the parameters of the main classifier.

The two losses are combined to form the total loss \(l\) as follows:
\begin{equation}
    l = {l_o} + \lambda {l_a}
    \label{eq8}
\end{equation}
where $\lambda$ is a hyperparameter. Note that the auxiliary loss branch is only used during training.

\section{Experimental results and discussion}
\label{Experimental Results and Discussion}
In this section, we validate our indoor scene recognition method through extensive comparative experiments. We begin by introducing the datasets and providing experimental details, followed by an analysis of critical parameters and their recommended values. Subsequently, we perform an ablation study of the proposed method. Finally, we present and thoroughly analyze the qualitative and quantitative experimental results.

\subsection{Datasets}
\textbf{MIT-67} Dataset \cite{quattoni2009recognizingr18} comprises 15,620 indoor scene images distributed across 67 categories. Each category contains a minimum of 100 images. According to \cite{quattoni2009recognizingr18}, we allocate 80 images per class for training and 20 for testing.

\textbf{SUN397} Dataset \cite{xiao2010sunr19} encompasses 397 scene classes, encompassing 175 indoor scenes and 220 outdoor scenes, each containing at least 100 RGB images. This study focuses on complex indoor scenes, using 175 indoor scene classes to evaluate the proposed method. Following the protocol in \cite{xiao2010sunr19}, we randomly select 50 images from each scene class for training and another 50 for testing.

\textbf{Places} Dataset \cite{zhou2017placesr20} is an extensive scene dataset, featuring approximately 1.8 million training images.  Each scene category within this dataset consists of 5000 images for training and 100 images for evaluation. For equal comparison with other methods \cite{zhou2021bormr14,miao2021objectr12} for indoor scene recognition, we use a simplified version of this dataset, i.e., Places\_7 and Places\_14. The former contains seven indoor scenes: Bathroom, Bedroom, Corridor, Dining Room, Kitchen, Living Room, and Office. The latter contains 14 indoor scenes: Balcony, Bedroom, Dining Room, Home Office, Kitchen, Living Room, Staircase, Bathroom, Closet, Garage, Home Theater, Laundromat, Playroom, and Wet Bar.

The specific splits of each dataset can be seen in Table \ref{tab_statistic}.

\subsection{Implementation details}
\subsubsection{Semantic segmentation}
In our experiment, semantic segmentation is performed using the Vision Transformer Adapter \cite{chen2022visionr21} trained on the ADE20K dataset \cite{zhou2019semanticr22}. The number of segmentable objects by this technique is shown in Table \ref{tab_statistic}. Given the inherent limitations of semantic segmentation precision, we use the Adaptive Confidence Filter \cite{song2023Srrmr10} to refine the semantic segmentation score tensor before generating the label map. For computational efficiency, we employ nearest neighbor interpolation to align the resolution of the label map with that of the feature map obtained from the backbone. It is worth noting that increasing the resolution might improve recognition accuracy, but this is not the focus of our work.

\begin{table}[!ht]
    \centering
    \caption{Dataset statistics, where the number of training images, test images, categories, segmentable objects, nodes of graph, and edges of graph for each dataset are listed below.}
    \begin{adjustbox}{width=\textwidth}
    \begin{tabular}{lcccccc}
    \hline
        Dataset & training images & test images & categories & segmentable objects & graph nodes & edges \\ \hline
        MIT-67 & 5630 & 1340 & 67  & $L=150$ & $n={w_1}{h_1}$ & $n^2$ \\ 
        SUN397 & 8750 & 8750 & 175  & $L=150$ & $n={w_1}{h_1}$ & $n^2$ \\ 
        Places\_7 & 35000 & 700 & 7  & $L=150$ & $n={w_1}{h_1}$ & $n^2$ \\ 
        Places\_14 & 70000 & 1400 & 14  & $L=150$ & $n={w_1}{h_1}$ & $n^2$ \\ \hline
    \end{tabular}
    \end{adjustbox}
    \label{tab_statistic}
\end{table}

\subsubsection{Hyperparameters}
The input image resolution of the backbone network is fixed at 224×224. During the training phase, we train our models for 30 epochs using the Adam optimizer \cite{kingma2014adamr23}. All parameters are randomly initialized following the Xavier initialization method \cite{glorot2010understandingr24}. The graph module is initialized with a learning rate of 0.001, which is reduced by a factor of 0.1 at the 10th, 15th, and 20th epochs. To preserve the feature extraction capability of the backbone, we set its learning rate to 0.00001 during the initial training phase and decrease it by a factor of 0.1 at the 10th epoch. We also apply a weight decay of 0.00001 to regularize the model. For the MIT-67 \cite{quattoni2009recognizingr18} and SUN397 \cite{xiao2010sunr19} datasets, the batch size is set to 32. For the Places \cite{zhou2017placesr20} dataset, we change the batch size to 64 due to its larger size and faster convergence. All experiments are conducted on a single NVIDIA 3090 GPU using the PyTorch \cite{paszke2019pytorchr33} and SAS-Net \cite{lopez2020semanticr9} framework. Note that the random seed was set to 304 in all training processes to ensure the reproducibility of all experiments.

We use ConvNext \cite{liu2022convnetr25} as our backbone network for feature extraction. To conserve computing resources, we initially conduct the ablation study with ConvNext\_Tiny pretrained on ImageNet 1k \cite{russakovsky2015imagenetr26}. Subsequently, we evaluate the performance of our proposed method with ConvNext\_Base pretrained on ImageNet 22k \cite{russakovsky2015imagenetr26}. Finally, for a fair comparison \cite{lin2022scener4,lopez2020semanticr9,song2023Srrmr10}, the standard 10-crop testing method is used for comparison with other methods.

\subsubsection{Baseline}
We constructed a baseline model that takes the output features of the backbone network as input, undergoes average pooling, and a fully connected layer for scene prediction. The performance achieved by this model will be used as a baseline for subsequent experiments.

\subsection{Design options}
In this section, we compare the design options for generating IODP as described in Section \ref{Inter-Object Discriminative Prototype}. For a fair comparison, we have temporarily set the value of $\lambda$ in Auxiliary loss to 1, and configured the hidden dimension ($d$) in GCN to be 768 for all experiments in this section.

\subsubsection{Object co-occurrence calculations}
\label{Object Co-occurrence Calculations}
This section presents a comparative analysis of the performance achieved by various methods for computing the object co-occurrence distribution. To maintain consistency, we employ the Coefficient of Variation ($cv$) to measure discriminative correlation between objects. The resulting experimental findings are presented in Table \ref{tab1}. In Table \ref{tab1}, "Non-independent distribution" signifies the computation of the co-occurrence distribution of two objects using the normal statistical method, i.e., $\frac{{{N_{{o_i},{o_j}}}}}{{{N_{{S_c}}}}}$ in Eq. \ref{eq1}. "Independent distribution" denotes that the respective distributions of the two objects are computed first. Then the co-occurrence distribution of the two objects is computed according to the assumption of independent distribution, i.e., $\frac{{{N_{{o_i}}} \times {N_{{o_j}}}}}{{N_{{S_c}}^2}}$ in Eq. \ref{eq1}. More details can be found in Eq. \ref{eq1}.

\begin{table}[!ht]
    \centering
    \caption{The effect of using different ways to compute the object co-occurrence distribution on classification accuracy (\%). The best result in each row is highlighted in bold.}
    \begin{adjustbox}{width=0.7\textwidth}
    \begin{tabular}{lcc}
    \hline
        Datasets & Non-independent distribution & Independent distribution  \\ \hline
        MIT-67 & 81.791 & \textbf{82.015}  \\ 
        SUN397 & 66.094 & \textbf{66.471}  \\ 
        Places\_7 & \textbf{92.714} & 92.229  \\ 
        Places\_14 & \textbf{88.571} & 88.5  \\ \hline
    \end{tabular}
    \end{adjustbox}
    \label{tab1}
\end{table}

Following the results in Table \ref{tab1}, we observe that on the Places\_7 and Places\_14 datasets, computing the co-occurrence probability distribution between two objects in the normal way ("Non-independent distribution") produces relatively better recognition performance. In contrast, on the MIT-67 and SUN397 datasets, computing the co-occurrence between two objects based on their independent distributions yields better performance. This finding aligns with the discussion in Section \ref{Inter-Object Discriminative Prototype}, possibly attributed to varying data quantities in the datasets.

In view of these empirical results, in all subsequent experiments, we employ different methods to compute object co-occurrence depending on the dataset size. For the MIT-67 and SUN397 datasets, with limited data, we utilize the "Independent distribution" for object co-occurrence calculation. For the Places\_7 and Places\_14 datasets, which contain ample data, we opt for the "Non-independent distribution" for object co-occurrence calculation.

\subsubsection{Inter-object discrimination correlation measurements}
\label{Inter-Object discrimination correlation measurements}
\begin{figure}[htbp]
    \centering
    \includegraphics[width=0.5\textwidth]{ 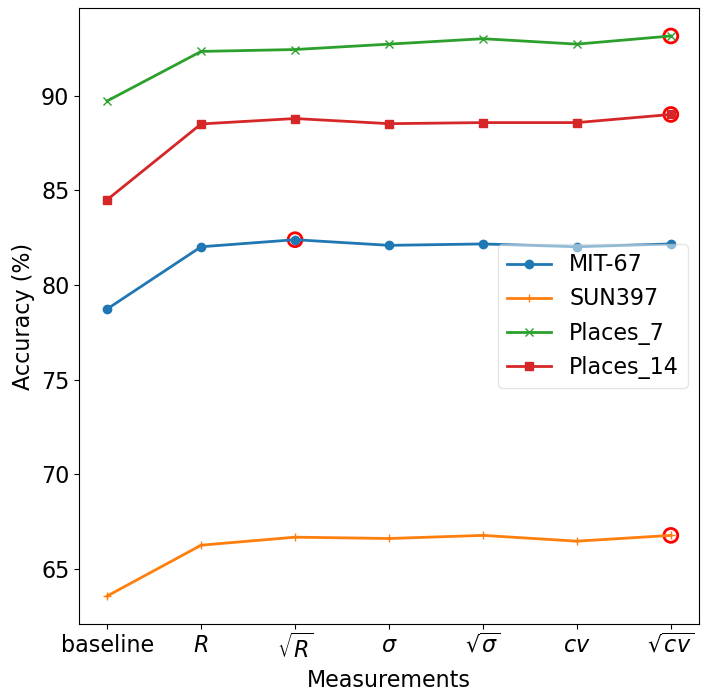}
    \caption{The effect of using different ways to measure the inter-object discriminative correlation on classification accuracy (\%). Note that the nodes circled by hollow bold circles are the highest accuracy points. $R$ represents the $Range$, $\sigma$ denotes the Standard deviation, and $cv$ represents the Coefficient of variation.}
    \label{Fig5}
\end{figure}

To determine the optimal way to quantify inter-object discriminative correlation, we conduct a comprehensive evaluation of various metrics and their respective variants, as outlined in Eq. \ref{eq4} and Eq. \ref{eq5}. The recognition results produced by these metrics are shown in Fig. \ref{Fig5}.

It is obvious that our approach significantly improves network performance over the baseline across all datasets, irrespective of the metric used to evaluate inter-object discriminative correlations. These results further demonstrate the robustness and generalization of using IODP to provide prior knowledge for scene recognition. Moreover, applying a square root transformation to the metric results typically leads to improved performance regardless of the metric employed. This is because the process of calculating object distributions ignores some detailed information in the scene (e.g., color, etc.), potentially resulting in sharper discriminative relationships between objects than reality. This can be effectively mitigated by square root transformation. Also, it is noteworthy that $\sqrt {cv} $ consistently yields optimal recognition results for most datasets. For consistency, we use $\sqrt {cv} $ to measure inter-object discriminative correlations in all subsequent experiments.

\subsection{Hyperparameter analysis}
Two hyperparameters, the hidden dimension $d$ in GCN and the $\lambda$ in Auxiliary loss, play pivotal roles in determining the performance of our method. To investigate the influence of these hyperparameters, we conduct several experiments using the ConvNext\_Tiny network.

\begin{figure}[htbp]
    \centering
    \includegraphics[width=0.5\textwidth]{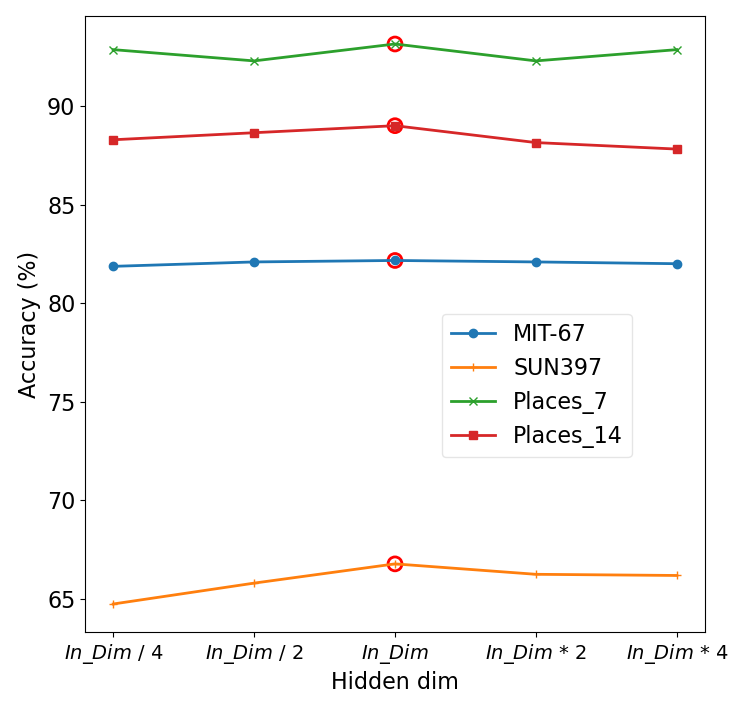}
    \caption{The effect of hidden dim $d$ in GCN. Note that the nodes circled by hollow bold circles are the highest accuracy points. $In\_Dim$ represents the channel dimension of the feature map extracted by the backbone (for ConvNext\_Tiny, $In\_Dim$ = 768; for ConvNext\_Base, $In\_Dim$ = 1024).}
    \label{Fig6}
\end{figure}

Initially, we perform experiments to assess the impact of different hidden dimensions ($d$) on multiple datasets. Note that the value of $\lambda$ is fixed to 1 to ensure a fair comparison. As shown in Fig. \ref{Fig6}, it can be observed that most datasets show a consistent trend, i.e., the recognition performance gradually improves with increasing hidden dimensions. However, a notable deviation arises when the hidden dimension reaches $In\_Dim$. At this point, the accuracy starts to decrease, possibly due to overfitting due to too high hidden dimension. In light of this finding, we opt to set the hidden dimension ($d$) equal to $In\_Dim$ to maintain flexibility in parameter tuning and mitigate the risk of overfitting. It is crucial to mention that $In\_Dim$ represents the channel dimension of the feature map extracted by the backbone network.

Apart from $d$, the choice of hyperparameter $\lambda$ in the Auxiliary loss is also crucial. Although incorporating the IODP improves the network's overall performance, it may negatively affect the backbone's discriminability. To address this concern, we introduce an auxiliary loss during training, which preserves the intrinsic discriminability of the backbone network. Table \ref{tab2} displays recognition results for various $\lambda$ values. It is evident that, for the MIT-67, Places\_7, and Places\_14 datasets, the optimal performance is achieved when $\lambda$ is set to 0.25. Conversely, SUN397 attains the highest accuracy at $\lambda$ = 0.5.

\begin{table}[htbp]
    \centering
    \caption{The influence of $\lambda$ on classification accuracy (\%) when combined with auxiliary loss. The best result of each row is marked in bold.}
    \begin{adjustbox}{width=0.6\textwidth}
    \begin{tabular}{cccccc}
    \hline
        $\lambda$ & 0 & 0.25 & 0.5 & 1 & 2 \\ \hline
        MIT-67 & 82.090 & \textbf{82.239} & 81.791 & 82.164 & 82.09  \\ 
        SUN397 & 66.741 & 66.753 & \textbf{66.929} & 66.776 & 66.506  \\ 
        Places\_7 & 92.286 & \textbf{93.143} & 92.143 & 93.143 & 93.086  \\ 
        Places\_14 & 88.357 & \textbf{89.071} & 88.5 & 89 & 88.714  \\ \hline
    \end{tabular}
    \end{adjustbox}
    \label{tab2}
\end{table}

Based on the conducted experiments, we determine the optimal hyperparameters for evaluating our method. The specific parameter values are as follows: \{$d = In\_Dim,\lambda {\text{ = 0}}{\text{.25}}$\} on MIT-67, Places\_7 and Places\_14, and \{$d = In\_Dim,\lambda {\text{ = 0}}{\text{.5}}$\} on SUN397. These parameter settings are set as the default configuration for subsequent experiments.

\subsection{Ablation study and evaluation}
We conduct a comprehensive ablation study of our approach using MIT-67 and Places\_14 datasets. We use ConvNext\_Tiny as the backbone network, and pass its features through a fully connected layer with average pooling to obtain the baseline results. We then analyze the effect of using IODP to provide prior knowledge to the trained network (only evaluation with IODP). Additionally, we explore the effect of auxiliary loss. Experimental results are presented in Table \ref{tab3}. Note that when only evaluation with IODP, we remove the hidden layer and activation function in GCN, i.e., changed $V^* = \eta ({D^{ - 1}}\widetilde AVW)$ to ${{V^ * } = {D^{ - 1}}\widetilde AV}$ in Eq. \ref{eq7}. In this way, the prior knowledge in IODP can be plug-and-played into the trained baseline models.

\begin{table}[!ht]
    \centering
    \caption{Ablation studies on MIT-67 and Places\_14. Classification accuracy (\%) is reported as the evaluation metric.}
    \begin{adjustbox}{width=0.7\textwidth}
    \begin{tabular}{cccccc}
    \hline
        Baseline & Train with IODP & Eval with IODP & Auxiliary loss & MIT-67 & Places\_14 \\ \hline
        \checkmark & - & - & - & 78.731 & 84.5 \\ 
        - & - & \checkmark & - & 79.478 & 87.357 \\ 
        - & \checkmark & \checkmark & - & 82.090 & 88.357 \\ 
        - & \checkmark & \checkmark & \checkmark & \textbf{82.239} & \textbf{89.071} \\ \hline
    \end{tabular}
    \end{adjustbox}
    \label{tab3}
\end{table}

In Table \ref{tab3}, the best results are highlighted in bold, with 3.508\% and 4.571\% improvement over baseline on the MIT-67 and Places\_14 datasets, respectively. When IODP provides prior knowledge to the trained baseline network (only eval with IODP), it consistently enhances the baseline's performance, with a notable improvement of 2.857\% on Places\_14. This outcome is achieved without training or adding parameters to the network, indicating the direct applicability of IODP to trained networks. This strongly demonstrates the significance of the discriminative prior knowledge within IODP and the feasibility of using it from a graph perspective. The network performance is further improved when IODP is applied during both the training and evaluation phases. This phenomenon can be attributed to the enhanced utilization of the prior knowledge of IODP, facilitated by hidden layers after training. Moreover, by introducing the auxiliary loss, the network performance is further improved. This result is consistent with the analysis in Section \ref{Scene Classifier and Auxiliary loss}, indicating that integrating IODP in training to boost feature discriminability potentially weakens the training of the backbone network parameters. The proposed auxiliary loss can guarantee adequate training of the backbone network parameters.

\subsection{Qualitative analysis of our method}
\begin{figure}[htbp]
    \centering
    \includegraphics[width=0.9\textwidth]{ 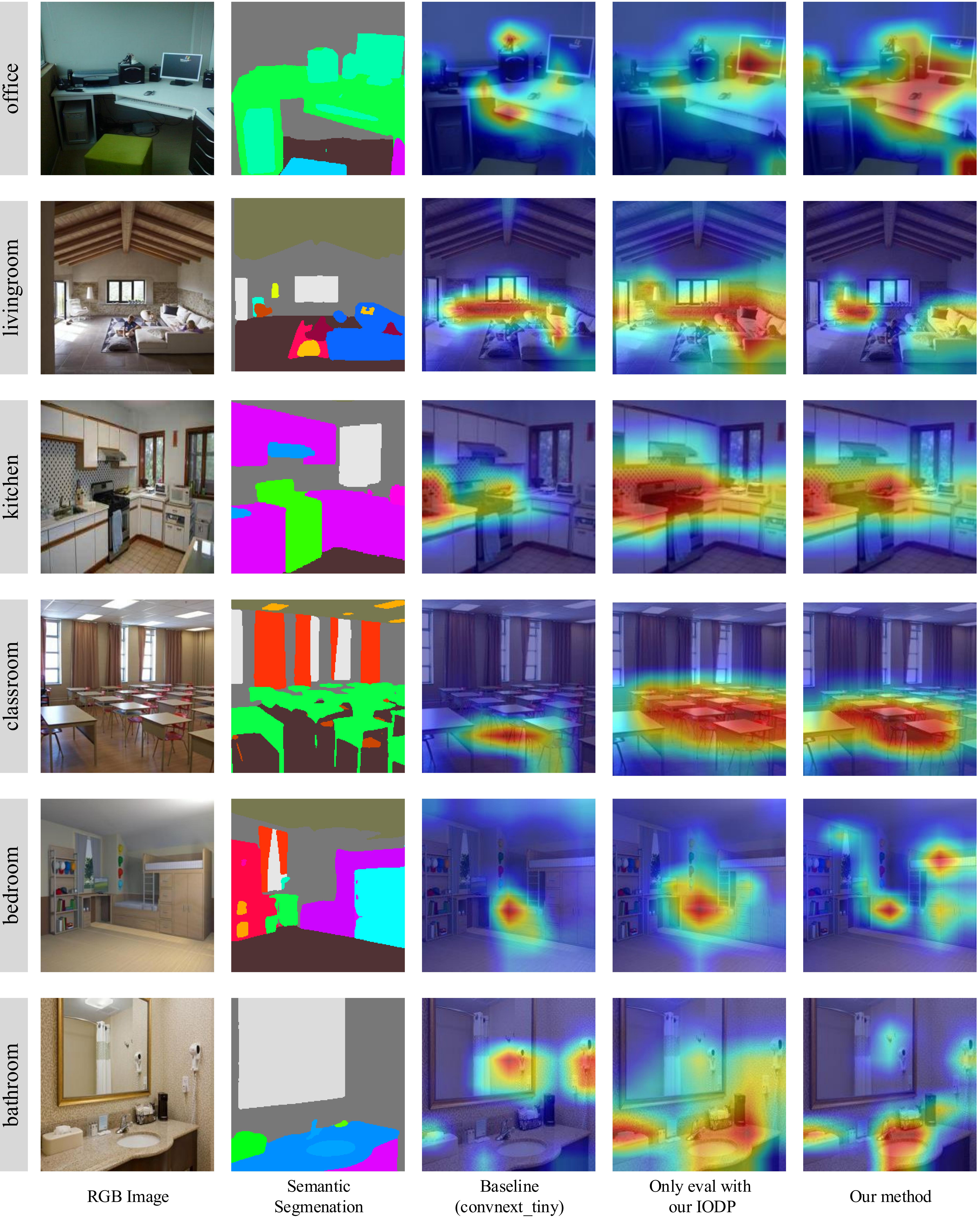}
    \caption{Qualitative visualization results. The first two columns depict RGB images from the MIT-67 \cite{quattoni2009recognizingr18} and Places \cite{zhou2017placesr20} validation sets alongside their semantic segmentation results. The next three columns showcase the Grad-CAM \cite{selvaraju2017gradr35} results generated via the feature extraction methods: baseline, only evaluation with our IODP, and our full method. Regions that are more emphasized by the network will appear more vividly red in the visualization outcomes.}
    \label{Fig7}
\end{figure}

To conduct a qualitative analysis of the advantages offered by our proposed method, we visualize the discriminative regions of each model using Grad-CAM \cite{selvaraju2017gradr35}. Fig. \ref{Fig7} displays the visualization results in the following sequence, from left to right: the original color image, the semantic segmentation results, the visualization of discriminative regions based on the baseline, the visualization using the IODP only during evaluation, and the visualization of the discriminative region based on our full DGN. The intensity of the red color indicates the network's focus on the corresponding region.

As demonstrated in Fig. \ref{Fig7}, 1) Compared to the visualization of the baseline results, the incorporation of the IODP enhances the network's performance (columns 4 and 5) by directing the model's attention to discriminative object regions. 2) When the IODP is applied only during evaluation (column 4), it yields favorable results with some adjustment to the baseline network's focus. However, using IODP for networks only in evaluation may lead to over-generalization, widening the network's focal area and introducing noise that degrades its discrimination. This limitation arises from the fact that IODP solely considers inter-object discriminability and does not have the ability to adjust intra-scene features based on factors such as color. In contrast, incorporating the IODP in both training and evaluation enables the refinement of network parameters and the discriminative graph through mechanisms such as hidden layers, thereby enhancing the rationality and effectiveness of the resulting interest regions.

\subsection{Limitations of the proposed method}
\label{Limitations}
According to the reported results, the proposed IODP is effective in assisting scene recognition. However, the performance of our method could be constrained or even hindered when the discriminative objects within scenes exceed the detection range of the employed semantic segmentation technique.

To illustrate this, we can refer to the last row in Fig. \ref{Fig7}, where the "hair dryer" serves as a discriminative object within the depicted "bathroom." The attention to "hair dryer" is evident in the visualization result corresponding to the baseline (third column). However, the semantic segmentation technique is unable to recognize the "hair dryer," resulting in a substantial decrease in the network's attention on the "hair dryer" region after integrating the IODP (fourth column). Fortunately, our ultimate method (fifth column) addresses this problem by adaptively adjusting the network parameters for our graph during training, thus partially maintaining the attention on the "hair dryer" while relying on other discriminative objects to accurately classify the scene category. Nevertheless, assuming that semantic segmentation fails to identify discriminative objects in a scene, especially when the number of discriminative objects is limited, our approach may ultimately fail to accurately classify the scene.

\subsection{Evaluation using ConvNext\_Base pretrained on ImageNet 22k}

\begin{table}[htbp]
    \centering
    \caption{Validation of our method on several datasets based on different models. The last row(*) indicates the evaluation using the 10-crop testing method \cite{lopez2020semanticr9}.}
    \begin{adjustbox}{width=0.7\textwidth}
    \begin{tabular}{c|ccccc}
    \hline
        Backbone & Method & MIT-67 & SUN397 & Places\_7 & Places\_14  \\ \hline
        \multirow{2}*{ConvNext\_Tiny} & baseline & 78.731 & 63.576 & 89.714 & 84.5  \\ 
        ~ & Ours  & 82.239 & 66.929 & 93.143 & 89.071  \\ \hline
        \multirow{3}*{ConvNext\_Base} & baseline & 87.388 & 75.424 & 93.857 & 88.957  \\ 
        ~ & Ours  & 89.254 & 78.671 & 94.286 & 89.914  \\ 
        ~ & Ours* & 90.373 & 79.765 & 94.286 & 89.914  \\ \hline
    \end{tabular}
    \end{adjustbox}
    \label{tab4}
\end{table}

Through our conducted experiments, we employ ConvNext\_Tiny as the backbone network to determine the design options and hyperparameters of the proposed method. Our ablation study and visualization analysis effectively demonstrate the significant support provided by our proposed method for scene recognition tasks. To further validate the effectiveness of our approach, we adopt ConvNext\_Base, which offers stronger feature extraction capabilities, as the backbone network to assess its performance on the  MIT-67, SUN397, Places\_7, and Places\_14 datasets. The experimental results are presented in Table \ref{tab4}. It can be observed that ConvNext\_Base, with its powerful feature extraction ability, achieves higher recognition accuracy on these scene datasets. Nevertheless, our proposed method consistently improves the accuracy, providing additional evidence of the utility of the IODP and DGN for scene recognition. Finally, in the last row of Table \ref{tab4}, we evaluate our DGN using the standard 10-crop testing method \cite{lopez2020semanticr9}. Since we use a version of the Places \cite{zhou2017placesr20} dataset resized to the network input, the 10-crop method has no impact on the evaluation of the Places dataset.

\subsection{Comparison with state-of-the-art methods}
In this section, we compare our proposed DGN with state-of-the-art methods, ranging from general deep architectures to methods that use object information to assist in scene recognition. These comparisons are conducted on publicly available datasets: MIT-67, SUN397, Places\_7, and Place\_14. As presented in Tables \ref{tab5} and \ref{tab6}, we extract the results for all methods directly from their respective papers.

\begin{table}[!ht]
    \centering
    \caption{State-of-the-art results on Places\_14 and Places\_7 datasets.}
    \begin{adjustbox}{width=0.55\textwidth}
    \begin{tabular}{cccc}
    \hline
        Approaches & Input Size & Places\_14 & Places\_7 \\ \hline
        Word2Vec \cite{chen2019scener36}   & 224 $\times$ 224  & 83.7 & -  \\ 
        Deduce \cite{pal2019deducer37}   & 224 $\times$ 224  & - & 88.1  \\ 
        BORM-Net \cite{zhou2021bormr14}   & 224 $\times$ 224  & 85.8 & 90.1  \\ 
        OTS-Net \cite{miao2021objectr12}   & 224 $\times$ 224  & 85.9 & 90.1  \\ 
        CSRRM \cite{song2023Srrmr10}   & 224 $\times$ 224  & 88.714 & 93.429  \\ 
        AGCN \cite{zhou2023attentionalr38}   & 224 $\times$ 224  & 86.0 & 91.7  \\ 
        Our DGN  & 224 $\times$ 224  & \textbf{89.914} & \textbf{94.286 } \\ \hline
    \end{tabular}
    \end{adjustbox}
    \label{tab5}
\end{table}

\begin{table}[!ht]
    \centering
    \caption{State-of-the-art results on MIT-67 and SUN397 datasets.}
    \begin{adjustbox}{width=0.55\textwidth}
    \begin{tabular}{cccc}
    \hline
        Approaches  & Input Size  & MIT-67  & SUN397  \\ \hline
        Adi-Red \cite{zhao2018volcanor5}  & 224 $\times$ 224  & -  & 73.59  \\ 
        LGN \cite{chen2020scener3}   & 448 $\times$ 448  & 88.06  & 74.06  \\ 
        Scene Essence \cite{sceneessencer11}   & 224 $\times$ 224  & 83.92  & 68.31  \\ 
        MRNet \cite{lin2022scener4}   & 448 $\times$ 448  & 88.08  & 73.98  \\ 
        SDO \cite{cheng2018scener7}   & 224 $\times$ 224  & 86.76  & 73.41  \\ 
        SAS-Net \cite{lopez2020semanticr9}   & 224 $\times$ 224  & 87.1  & 74.04  \\ 
        CCF \cite{sitaula2021contentr40}   & 512 $\times$ 512  & 87.3  & -  \\ 
        ARG-Net \cite{zeng2020amorphousr15}   & 448 $\times$ 448  & 88.13  & 75.02  \\ 
        FCT \cite{xie2022fctr39}   & 448 $\times$ 448  & \textcolor{green}{90.75}  & 77.50  \\ 
        CSSRM \cite{song2023Srrmr10}   & 224 $\times$ 224  & 88.731  & -  \\ 
        Our DGN  & 224 $\times$ 224  & \textbf{90.373}  & \textbf{79.765}  \\ \hline
    \end{tabular}
    \end{adjustbox}
    \label{tab6}
\end{table}

The results in Tables \ref{tab5} and \ref{tab6} demonstrate that our method outperforms most existing methods. First, our method achieves superior accuracy compared to methods \cite{chen2020scener3,lin2022scener4,zhao2018volcanor5,zhou2023attentionalr38,sitaula2021contentr40} that employ convolutional responses to select features for scene recognition. This demonstrates the potential of using specific object information within the scene to enhance feature discriminability. The focus then shifts to object-assisted methods. Approaches \cite{cheng2018scener7,lopez2020semanticr9,song2023Srrmr10,sceneessencer11,zhou2021bormr14,pal2019deducer37} use multi-branch architectures to separately process scene images and object information, subsequently combining object features with scene features through concatenation or summation to obtain a scene representation. In contrast, our work employs a single-branch architecture that adjusts scene features using object knowledge from a theoretical perspective. The proposed approach not only enhances the recognition performance, but also alleviates the computational burden that multi-branch architectures may impose. Despite \cite{xie2022fctr39} exhibiting 0.377\% higher accuracy than ours on MIT-67, our method is more lightweight due to the larger input size and multi-branch structure of \cite{xie2022fctr39}. In addition, our method outperforms \cite{xie2022fctr39} by 2.265\% on SUN397, indicating its greater generalizability. Meanwhile, in comparison to methods \cite{sceneessencer11,zeng2020amorphousr15,zhou2023attentionalr38} that also utilize graph convolution to support scene recognition, our approach integrates discriminative prior knowledge into the graph, resulting in superior performance. This further highlights the potential of exploring inter-object discriminative prior knowledge.

It is noteworthy that methods \cite{miao2021objectr12,zeng2020amorphousr15} also use a single-branch architecture. Still, they only leverage the pixel locations of the object information and do not explore the discriminative knowledge embedded in the object information, resulting in suboptimal performance. Furthermore, approaches \cite{cheng2018scener7,zhou2021bormr14} apply a comparable strategy to extract discriminative correlations between objects. However, their separate branch for processing object information is limited, as object knowledge is solely used in that branch, resulting in limited utilization of object information. This also confirms the significance of inter-object discriminative knowledge for scene recognition, and our DGN can effectively utilize this knowledge to guide scene features toward more discriminative image representations.

\section{Conclusions}
\label{conclusion}
In this paper, we propose capturing object-scene discriminative relationships from a probabilistic perspective, and subsequently transforming them into Inter-Object Discriminative Prototype (IODP). IODP contains abundant object-object discriminative knowledge, which can be used to enhance the performance of scene recognition networks. Specifically, we frame the discriminative assistance process as a Discriminative Graph Network (DGN). DGN efficiently translates the inter-object discriminative knowledge from IODP into pixel-level correlations within scene features. Inter-object discriminative knowledge and scene features are transformed into discriminative image representations through graph convolution and mapping operations (GCN). Through qualitative and quantitative analyses on several public datasets, including MIT-67, SUN397, Places\_7, and Places\_14, we validate the significance of inter-object discriminative knowledge in scene recognition, and demonstrate the effectiveness and generalizability of using this knowledge for scene feature improvement through graph theory.

However, as mentioned in Section \ref{Limitations}, the proposed approach is limited by the number of objects that can be recognized by semantic segmentation techniques. Therefore, in future research, we intend to explore automatic scene object segmentation via unsupervised or semi-supervised methods, overcoming the limitations imposed by semantic segmentation techniques. SAGN \cite{yang2023sagnr6} has already initiated preliminary attempts towards this idea, which is a promising direction that deserves in-depth research.

\section*{Acknowledgment}
This work was jointly supported by the Key Development Program for Basic Research of Shandong Province under Grant ZR2019ZD07, the National Natural Science Foundation of China-Regional Innovation Development Joint Fund Project under Grant U21A20486, the Fundamental Research Funds for the Central Universities under Grant 2022JC011, the Natural Science Youth Foundation of Shandong Province under Grant ZR2023QF055.



\bibliographystyle{elsarticle-num} 

\begin{thebibliography}{00}


\bibitem{xie2020scener1}
L.~Xie, F.~Lee, L.~Liu, K.~Kotani, Q.~Chen, Scene recognition: A comprehensive survey, Pattern Recognition 102 (2020) 107205.

\bibitem{yuan2019acmr2}
Y.~Yuan, Z.~Xiong, Q.~Wang, Acm: Adaptive cross-modal graph convolutional neural networks for rgb-d scene recognition, in: Proceedings of the AAAI conference on artificial intelligence, Vol.~33, 2019, pp. 9176--9184.

\bibitem{chen2020scener3}
G.~Chen, X.~Song, H.~Zeng, S.~Jiang, Scene recognition with prototype-agnostic scene layout, IEEE Transactions on Image Processing 29 (2020) 5877--5888.

\bibitem{zhao2018volcanor5}
Z.~Zhao, M.~Larson, From volcano to toyshop: Adaptive discriminative region discovery for scene recognition, in: Proceedings of the 26th ACM international conference on Multimedia, 2018, pp. 1760--1768.

\bibitem{lin2022scener4}
C.~Lin, F.~Lee, L.~Xie, J.~Cai, H.~Chen, L.~Liu, Q.~Chen, Scene recognition using multiple representation network, Applied Soft Computing 118 (2022) 108530.

\bibitem{lopez2020semanticr9}
A.~L{\'o}pez-Cifuentes, M.~Escudero-Vinolo, J.~Besc{\'o}s, {\'A}.~Garc{\'\i}a-Mart{\'\i}n, Semantic-aware scene recognition, Pattern Recognition 102 (2020) 107256.

\bibitem{song2023Srrmr10}
C.~Song, X.~Ma, Srrm: Semantic region relation model for indoor scene recognition, in: 2023 International Joint Conference on Neural Networks (IJCNN), 2023, pp. 01--08.

\bibitem{sceneessencer11}
J.~Qiu, Y.~Yang, X.~Wang, D.~Tao, Scene essence, in: 2021 IEEE/CVF Conference on Computer Vision and Pattern Recognition (CVPR), 2021, pp. 8318--8329.

\bibitem{sitaula2021contentr40}
C.~Sitaula, S.~Aryal, Y.~Xiang, A.~Basnet, X.~Lu, Content and context features for scene image representation, Knowledge-Based Systems 232 (2021) 107470.

\bibitem{cheng2018scener7}
X.~Cheng, J.~Lu, J.~Feng, B.~Yuan, J.~Zhou, Scene recognition with objectness, Pattern Recognition 74 (2018) 474--487.

\bibitem{pereira2021deepr13}
R.~Pereira, L.~Garrote, T.~Barros, A.~Lopes, U.~J. Nunes, A deep learning-based indoor scene classification approach enhanced with inter-object distance semantic features, in: 2021 IEEE/RSJ International Conference on Intelligent Robots and Systems (IROS), IEEE, 2021, pp. 32--38.

\bibitem{zhou2021bormr14}
L.~Zhou, J.~Cen, X.~Wang, Z.~Sun, T.~L. Lam, Y.~Xu, Borm: Bayesian object relation model for indoor scene recognition, in: 2021 IEEE/RSJ International Conference on Intelligent Robots and Systems (IROS), IEEE, 2021, pp. 39--46.

\bibitem{choe2021indoorr16}
S.~Choe, H.~Seong, E.~Kim, Indoor place category recognition for a cleaning robot by fusing a probabilistic approach and deep learning, IEEE Transactions on Cybernetics 52~(8) (2021) 7265--7276.

\bibitem{miao2021objectr12}
B.~Miao, L.~Zhou, A.~S. Mian, T.~L. Lam, Y.~Xu, Object-to-scene: Learning to transfer object knowledge to indoor scene recognition, in: 2021 IEEE/RSJ International Conference on Intelligent Robots and Systems (IROS), IEEE, 2021, pp. 2069--2075.

\bibitem{zeng2020amorphousr15}
H.~Zeng, X.~Song, G.~Chen, S.~Jiang, Amorphous region context modeling for scene recognition, IEEE Transactions on Multimedia 24 (2022) 141--151.

\bibitem{yang2023sagnr6}
Y.~Yang, X.~Tang, Y.-M. Cheung, X.~Zhang, L.~Jiao, Sagn: Semantic-aware graph network for remote sensing scene classification, IEEE Transactions on Image Processing 32 (2023) 1011--1025.

\bibitem{kipf2016semir17}
T.~N. Kipf, M.~Welling, Semi-supervised classification with graph convolutional networks, in: ICLR, 2017.

\bibitem{quattoni2009recognizingr18}
A.~Quattoni, A.~Torralba, Recognizing indoor scenes, in: 2009 IEEE conference on computer vision and pattern recognition, IEEE, 2009, pp. 413--420.

\bibitem{xiao2010sunr19}
J.~Xiao, J.~Hays, K.~A. Ehinger, A.~Oliva, A.~Torralba, Sun database: Large-scale scene recognition from abbey to zoo, in: 2010 IEEE computer society conference on computer vision and pattern recognition, IEEE, 2010, pp. 3485--3492.

\bibitem{zhou2017placesr20}
B.~Zhou, A.~Lapedriza, A.~Khosla, A.~Oliva, A.~Torralba, Places: A 10 million image database for scene recognition, IEEE transactions on pattern analysis and machine intelligence 40~(6) (2017) 1452--1464.

\bibitem{liu2022convnetr25}
Z.~Liu, H.~Mao, C.-Y. Wu, C.~Feichtenhofer, T.~Darrell, S.~Xie, A convnet for the 2020s, in: Proceedings of the IEEE/CVF conference on computer vision and pattern recognition, 2022, pp. 11976--11986.

\bibitem{liu2019novelr31}
S.~Liu, G.~Tian, Y.~Xu, A novel scene classification model combining resnet based transfer learning and data augmentation with a filter, Neurocomputing 338 (2019) 191--206.

\bibitem{leng2023multitask}
Y.~Leng, J.~Zhuang, J.~Pan, C.~Sun, Multitask learning for acoustic scene classification with topic-based soft labels and a mutual attention mechanism, Knowledge-Based Systems 268 (2023) 110460.

\bibitem{yuan2022scaler32}
X.~Yuan, Z.~Qiao, A.~Meyarian, Scale attentive network for scene recognition, Neurocomputing 492 (2022) 612--623.

\bibitem{zhou2016learningr34}
B.~Zhou, A.~Khosla, A.~Lapedriza, A.~Oliva, A.~Torralba, Learning deep features for discriminative localization, in: Proceedings of the IEEE conference on computer vision and pattern recognition, 2016, pp. 2921--2929.

\bibitem{pal2019deducer37}
A.~Pal, C.~Nieto-Granda, H.~I. Christensen, Deduce: Diverse scene detection methods in unseen challenging environments, in: 2019 IEEE/RSJ International Conference on Intelligent Robots and Systems (IROS), IEEE, 2019, pp. 4198--4204.

\bibitem{tian2022skeletonr28}
H.~Tian, X.~Ma, H.~Wu, Y.~Li, Skeleton-based abnormal gait recognition with spatio-temporal attention enhanced gait-structural graph convolutional networks, Neurocomputing 473 (2022) 116--126.

\bibitem{zhou2023static}
H.~Zhou, X.~Yang, M.~Fan, H.~Huang, D.~Ren, H.~Xia, Static-dynamic global graph representation for pedestrian trajectory prediction, Knowledge-Based Systems 277 (2023) 110775.

\bibitem{tian2023skeletonr29}
H.~Tian, X.~Ma, X.~Li, Y.~Li, Skeleton-based action recognition with select-assemble-normalize graph convolutional networks, IEEE Transactions on Multimedia (2023).

\bibitem{liu2023enhancingr30}
H.~Liu, Y.~Wu, Q.~Li, W.~Lu, X.~Li, J.~Wei, X.~Liu, J.~Feng, Enhancing aspect-based sentiment analysis using a dual-gated graph convolutional network via contextual affective knowledge, Neurocomputing 553 (2023) 126526.

\bibitem{zhou2023attentionalr38}
L.~Zhou, Y.~Zhou, X.~Qi, J.~Hu, T.~L. Lam, Y.~Xu, Attentional graph convolutional network for structure-aware audiovisual scene classification, IEEE Transactions on Instrumentation and Measurement 72 (2023) 1--15.

\bibitem{zeng2023naturalScene}
Z.~Zeng, X.~Wang, W.~Li, Y.~Ye, Two-stage natural scene image classification with noise discovering and label-correlation mining, Knowledge-Based Systems 260 (2023) 110137.

\bibitem{chen2022visionr21}
Z.~Chen, Y.~Duan, W.~Wang, J.~He, T.~Lu, J.~Dai, Y.~Qiao, Vision transformer adapter for dense predictions, in: ICLR, 2023.

\bibitem{zhou2019semanticr22}
B.~Zhou, H.~Zhao, X.~Puig, T.~Xiao, S.~Fidler, A.~Barriuso, A.~Torralba, Semantic understanding of scenes through the ade20k dataset, International Journal of Computer Vision 127 (2019) 302--321.

\bibitem{kingma2014adamr23}
D.~P. Kingma, J.~Ba, Adam: A method for stochastic optimization, in: ICLR, 2015.

\bibitem{glorot2010understandingr24}
X.~Glorot, Y.~Bengio, Understanding the difficulty of training deep feedforward neural networks, in: Proceedings of the thirteenth international conference on artificial intelligence and statistics, JMLR Workshop and Conference Proceedings, 2010, pp. 249--256.

\bibitem{paszke2019pytorchr33}
A.~Paszke, S.~Gross, F.~Massa, A.~Lerer, J.~Bradbury, G.~Chanan, T.~Killeen, Z.~Lin, N.~Gimelshein, L.~Antiga, et~al., Pytorch: An imperative style, high-performance deep learning library, Advances in neural information processing systems 32 (2019).

\bibitem{russakovsky2015imagenetr26}
O.~Russakovsky, J.~Deng, H.~Su, J.~Krause, S.~Satheesh, S.~Ma, Z.~Huang, A.~Karpathy, A.~Khosla, M.~Bernstein, et~al., Imagenet large scale visual recognition challenge, International journal of computer vision 115 (2015) 211--252.

\bibitem{selvaraju2017gradr35}
R.~R. Selvaraju, M.~Cogswell, A.~Das, R.~Vedantam, D.~Parikh, D.~Batra, Grad-cam: Visual explanations from deep networks via gradient-based localization, in: Proceedings of the IEEE international conference on computer vision, 2017, pp. 618--626.

\bibitem{chen2019scener36}
B.~X. Chen, R.~Sahdev, D.~Wu, X.~Zhao, M.~Papagelis, J.~K. Tsotsos, Scene classification in indoor environments for robots using context based word embeddings, in: 2018 International Conference on Robotics and Automation (ICRA) Workshop, 2018.

\bibitem{xie2022fctr39}
Y.~Xie, J.~Yan, L.~Kang, Y.~Guo, J.~Zhang, X.~Luan, Fct: fusing cnn and transformer for scene classification, International Journal of Multimedia Information Retrieval 11~(4) (2022) 611--618.

\end{thebibliography}


\end{document}